\documentclass[sigconf]{acmart}

\usepackage{natbib}
\usepackage{graphicx}
\usepackage{multirow}
\usepackage{amsmath}
\usepackage{booktabs}
\usepackage{subfigure}
\usepackage{epstopdf}
\usepackage{comment}
\usepackage{bbm}
\usepackage{bbold}
\usepackage{url}
\usepackage{xcolor}

\usepackage[T1]{fontenc}
\usepackage{aecompl}
\usepackage{balance} 

\begin{document}
\fancyhead{}

\copyrightyear{2022} 
\acmYear{2022} 
\setcopyright{acmlicensed}\acmConference[WSDM '22]{Proceedings of the Fifteenth ACM International Conference on Web Search and Data Mining}{February 21--25, 2022}{Tempe, AZ, USA}
\acmBooktitle{Proceedings of the Fifteenth ACM International Conference on Web Search and Data Mining (WSDM '22), February 21--25, 2022, Tempe, AZ, USA}
\acmPrice{15.00}
\acmDOI{10.1145/3488560.3498409}
\acmISBN{978-1-4503-9132-0/22/02}

\title{A Simple but Effective Bidirectional Framework \\ for Relational Triple Extraction}


\author{Feiliang Ren}
\authornote{Corresponding author, {renfeiliang@cse.neu.edu.cn}.}
\authornote{Both authors contribute equally to this research and are listed randomly.}
\orcid{0000-0001-6824-1191}
\affiliation{%
	\institution{Northeastern University}
	\city{Shenyang}
	\country{China}
}

\author{Longhui Zhang}
\affiliation{%
	\institution{Northeastern University}
	\city{Shenyang}
	\country{China}
}
\authornotemark[2]

\author{Xiaofeng Zhao}
\affiliation{%
	\institution{Northeastern University}
	\city{Shenyang}
	\country{China}
}

\author{Shujuan Yin}
\affiliation{%
	\institution{Northeastern University}
	\city{Shenyang}
	\country{China}
}

\author{Shilei Liu}
\affiliation{%
	\institution{Northeastern University}
	\city{Shenyang}
	\country{China}
}

\author{Bochao Li}
\affiliation{%
	\institution{Northeastern University}
	\city{Shenyang}
	\country{China}
}

\renewcommand{\shortauthors}{F. Ren, et al.}




\begin{abstract}
Tagging based relational triple extraction methods are attracting growing research attention recently. However, most of these methods  take a unidirectional extraction framework that first extracts all subjects and then extracts objects and relations simultaneously based on the subjects extracted. This framework has an obvious deficiency that it is too sensitive to the extraction results of subjects.  To overcome this deficiency, we propose a bidirectional extraction framework based method that extracts triples based on the entity pairs extracted from two complementary directions. Concretely, we first extract all possible subject-object pairs from two paralleled  directions. These two extraction directions are connected by a shared encoder component, thus the extraction features from one direction can flow to another direction and vice versa. By this way, the extractions of two directions can boost and complement each other. Next, we assign all possible relations for each entity pair by a biaffine model. During training, we observe that the share structure will lead to a convergence rate inconsistency issue which is harmful to performance. So we propose a share-aware learning mechanism to address it. We evaluate the proposed model on multiple benchmark datasets. Extensive experimental results show that the proposed model is very effective and it achieves state-of-the-art results on all of these datasets. Moreover, experiments show that both the proposed bidirectional extraction framework and the share-aware learning mechanism have good adaptability and can be used to improve the performance of other tagging based methods. The source code of our work is available at: https://github.com/neukg/BiRTE.
\end{abstract}




\maketitle

\section{Introduction}
\label{section:Introduction}
The task of relational triple extraction (RTE for short)  is to extract triples from unstructured  natural language text (often sentences). These  relational triples  store factual knowledge in the form of (\emph{subject, relation, object}), where both \emph{subject} and \emph{object} are entities and they are  connected semantically by \emph{relation}. For example, a triple (\emph{Washington, capital\_of, the United States}) expresses the knowledge that ``\emph{Washington is the capital of the United States}''. Nowadays,  RTE are attracting more and more research interest  due to its importance for many downstream applications like  automatic knowledge graph  construction, and many novel RTE methods have been proposed. 

Early RTE methods ~\cite{zelenko2003kernel,zhou2005exploring,chan2011exploiting} often use a pipeline based extraction framework that    recognizes all  entities in the input text first, and then  predicts the relations for all the combinations of entity pairs. These methods are flexible for they can make full use of  existing achievements in the research domains of both  name entity recognition  and relation classification. But they  have following two fatal deficiencies. First, they ignore the correlations between  entity recognition and relation prediction. Second, they suffer from the error propagation issue greatly. Thus more and more researchers begin to explore a kind of  joint extraction methods that extracts entities and relations simultaneously in an end-to-end way,  and lots of novel joint extraction methods have been proposed~\cite{zheng2017joint,bekoulis2018joint,fu2019graphrel,eberts2019span,yu2019joint,nayak2020effective,yuan2020a,zeng2020copymtl,wei-etal-2020-novel,wang-etal-2020-tplinker}.  

Among these joint extraction methods, a kind of tagging based methods~\cite{zheng2017joint,yu2019joint,wei-etal-2020-novel}  show great superiority in both the  performance and the ability of extracting triples from following two kinds of complex sentences. The first kind is the sentences  that contain overlapping  triples (a single entity or an entity pair participates in multiple relational triples of the same sentence~\cite{zeng2018Extracting}). The second kind is the sentences that   contains multiple triples. Existing tagging based methods often decompose the whole RTE task into two tagging based subtasks:  the first one recognizes all subjects and the second one recognizes all objects and relations simultaneously. For convenience, we call them as unidirectional extraction framework based  methods. 
Despite the great success, they are far from their full potential because  they suffer from the following issue greatly: once the extraction of a subject is failed, the extraction of all  triples associated with this subject would  be failed  accordingly. Here we call an entity  as a ground entity if it is extracted firstly in a triple, and call the mentioned issue as \emph{ground entity extraction failure}.   Obviously, this issue is  harmful to the performance of RTE greatly. 


To address the mentioned issue,  we propose \emph{BiRTE}, a \emph{Bi}directional extraction framework based \emph{R}elational \emph{T}riple \emph{E}xtraction model. It follows the  tagging based extraction route  but takes both subjects and objects as  ground entities. Our method is mainly inspired by  following cognition:  if we take both subjects and objects as ground entities and extract triples from the directions of ``\emph{subject->object->relation}'' and ``\emph{object->subject->relation}'', then even if a ground entity is failed to be extracted in one direction, it is still possible to be extracted from another direction (but not as a ground entity),  thus its associated triples are still possible to be extracted from another direction accordingly. Thus, the   mentioned \emph{ground entity extraction failure} issue can be well addressed inherently. 

Based on above cognition, \emph{BiRTE} is designed as follows. First, it extracts \emph{subject-object} (\emph{s-o} for short) pairs  from the directions of  \emph{subject-to-object} (\emph{s2o} for short) and  \emph{object-to-subject} (\emph{o2s} for short). These two  extraction directions work in parallel but are connected by a shared encoder component, which makes the extraction features from one direction can be injected into the extraction features of another direction, and vice versa. This extraction structure brings an obvious advantage: the  extractions of  two  directions complement each other and their extraction results can be  validated each other. And such advantage is much helpful for the whole triple extraction since reliable \emph{s-o} pairs are the foundation of extracting accurate triples. With this bidirectional extraction framework, 
lots of \emph{s-o} pairs are extracted. Among them, there are also many noise pairs that do not possess any relations.  Thus a strong relation classification model is required. In \emph{BiRTE}, we use a biaffine model to assigns all possible relations for each \emph{s-o} pair. Given a \emph{s-o} pair, the biaffine model can mine deep interactions between the subject and the object, thus all the relations of this \emph{s-o} pair can be easily extracted. Besides, during  training, we  observe there would be a harmful \emph{convergence rate inconsistency issue} caused by the share structure. To overcome it, we  propose a \emph{share-aware} learning mechanism which assigns different learning rates for different modules. 

%

We evaluate \emph{BiRTE} on multiple   benchmark datasets. Extensive experiments show  it consistently outperforms existing best RTE methods  on all datasets, and achieves new state-of-the-art results. 

\section{Related Work}
At present, the joint extraction methods are becoming dominated in  RTE. 
According to the extraction routes taken, we roughly classify them  into following three main kinds. 

\textbf{Tagging Based Methods}  In this kind of methods, binary tagging sequences are often used to determine the start and end positions of entities, sometimes are also used  to  determine the relations between two entities too. 
For example, \cite{zheng2017joint} propose a  tagging based framework that converts the joint extraction task into a tagging problem to extract entities and their relations directly. Recently, researchers \cite{yu2019joint,wei-etal-2020-novel} begin to explore a unidirectional extraction framework based tagging methods: first extract all subjects, and then extract objects and relations simultaneously based on the subjects extracted. 
Especially, CasRel~\cite{wei-etal-2020-novel}, one of the most latest tagging based methods,  provides a fresh perspective for the RTE task: it  models relations as functions that map subjects to objects. Experiments show that CasRel not only  achieves very competitive results, but also has strong ability for extracting triples from sentences that contain overlapping triples or multiple triples. 

\textbf{Table Filling Based Methods} This kind of methods ~\cite{wang-etal-2020-tplinker,zhang-etal-2017-end,miwa-bansal-2016-end,gupta-etal-2016-table} would maintain a $l \times l$ table for each relation ($l$ is the number of tokens in an input sentence), and the items in  this table  usually  denotes the start and end positions of two entities (or even the types of these entities) that possess this specific relation.  So the RTE task in this kind of methods is converted into the task of filling these tables accurately and effectively. 

\textbf{Seq2Seq Based Methods}  This kind of methods often  view  a triple as a token sequence, and convert the RTE task into a  \emph{generation} task that generates a triple in some orders,  such as first generates a relation, then generates   entities, etc. For example,  
~\cite{DBLP:conf/aaai/NayakN20}   use an encoder-decoder architecture in their method. ~\cite{DBLP:conf/aaai/YeZDCTHC21} propose a contrastive triple extraction method with a generative transformer. Other representative work of this kind includes ~\cite{zeng2018Extracting,zeng2020copymtl,zeng2019Learning}.


Researchers also explore other extraction routes for RTE. For example, 
~\cite{DBLP:conf/naacl/ChenZHH21} propose a unified framework to extract explicit and implicit relational triples jointly. ~\cite{DBLP:conf/acl/TianJHL20} provide a revealing insight into RTE from a stereoscopic perspective. ~\cite{DBLP:conf/acl/ZhengWCYZZZQMZ20} propose a joint RTE framework based on potential
relation and global correspondence.

\begin{figure*}[t]
	\centering
	\includegraphics[width=0.90\textwidth]{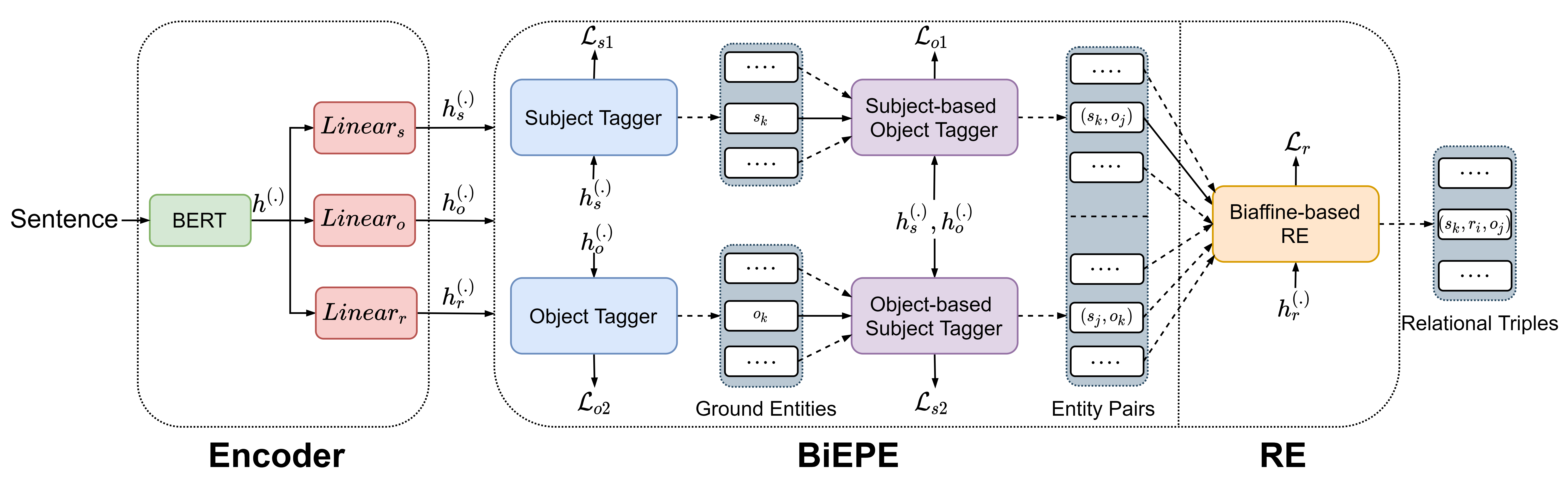}
	\caption{The Architecture of {BiRTE}. The modules with the same color have the similar inner structures. The solid lines represent the training process, and the dashed lines together with the solid lines  represent the inference phase. }
	\label{fig:model}
\end{figure*}

\section{Methodology}

The architecture of \emph{BiRTE} is shown in Figure~\ref{fig:model}, from which we can see that it consists of following three main components: a  BERT based \emph{Encoder} component, a \emph{Bi}directional \emph{E}ntity \emph{P}air \emph{E}xtraction  component (\emph{BiEPE} for short), and a biaffine based \emph{R}elation \emph{E}xtraction  component (\emph{RE} for short). 
During training, the modules in \emph{BiEPE}  and \emph{RE}   work in a multi-task learning manner. This brings an  advantage that each module can be trained with the ground-truth input, thus a more reliable model can be obtained.  But in the  inference phase, \emph{BiEPE} and  \emph{RE}   work in a sequential manner. 

\subsection{Encoder }
\label{sec:encoder}
Here we first use a pre-trained BERT (Cased) ~\cite{devlin2018bert}  model to generate an initial representation for each token (denoted as $\mathbf{h}^{(.)} \in \mathcal{R}^{d_h}$) in an input sentence. Then we generate three distinct token representation sequences  as a kind of context features for subjects, objects, and relations respectively. This is much different from most of existing state-of-the-art methods like CasRel~\cite{wei-etal-2020-novel}, TPLinker~\cite{wang-etal-2020-tplinker}, or PMEI~\cite{sun2021}: all of them use an unified feature for subjects, objects, and relations. But we think that different kinds of  items in triples have their own characteristics. Thus, it would be  more reasonable to use different features for them.   
Concretely, we denote the  $i$-th token representations of these three contextual features  as $\mathbf{h}_{s}^{i}$ , $\mathbf{h}_{o}^{i}$ , and $\mathbf{h}_{r}^i$ respectively, and they  are computed with Eq.\eqref{eq:token_rep}. 
\begin{equation}
\begin{aligned}
&\mathbf{h}_{s}^i = \mathbf{W}_{s} \mathbf{h}^i + \mathbf{b}_{s} \\&\mathbf{h}_{o}^i = \mathbf{W}_{o} \mathbf{h}^i + \mathbf{b}_{o}  \\ &\mathbf{h}_{r}^i = \mathbf{W}_{r} \mathbf{h}^i + \mathbf{b}_{r} \label{eq:token_rep}
\end{aligned}
\end{equation}
where $\mathbf{W}_{(.)} \in \mathcal{R}^{d_h \times d_h}$ is a trainable matrix,  $\mathbf{b}_{(.)} \in \mathcal{R}^{d_h} $ is a bias vector, and $d_h$ is the dimension.

Besides, considering the subject and object in a triple are highly correlated, the features from one entity would be helpful for the extraction of another entity. So  we add the object token representation sequence's \emph{CLS} vector
(denoted as $\mathbf{h}_{o}^{cls}$) to  $\mathbf{h}_{s}^i$  to enhance the representation ability of the subject's context features.  Similar operation is also performed on the object part, as  shown in  Eq.\eqref{eq:cls_add}.

\begin{equation}
\begin{aligned}
& \mathbf{h}_{s}^i = \mathbf{h}_{s}^i + \mathbf{h}_{o}^{cls}
\\
& \mathbf{h}_{o}^i = \mathbf{h}_{o}^i + \mathbf{h}_{s}^{cls} \label{eq:cls_add}
\end{aligned}
\end{equation}


\subsection{BiEPE }

\emph{BiEPE} has a bidirectional framework that extracts \emph{s-o} pairs from following two directions: (i) a \emph{s2o} direction that  first extracts subjects  and then extracts objects conditioned on the  subjects extracted,  and (ii) a reversed \emph{o2s} direction that first extracts objects and then extracts subjects. These two directions' extractions share the    \emph{Encoder} component. The inner structures of two directions are  similar, so here we only introduce the direction of \emph{s2o} for space saving.






\noindent \textbf{Subject Tagger} is a binary tagging based module  that aims to extract all subjects from an input sentence. For each token in the input sentence, two probabilities are assigned to denote the possibilities of it being the start token and end token of a subject. Specifically, these two probabilities   are computed with Eq.\eqref{eq:Subject Tagger}.
\begin{equation}
\begin{aligned}
&p_{s}^{i,start} = \sigma\left(\mathbf W_s^{start}\mathbf{h}_{s}^{i}+\mathbf b_s^{start}\right)\\
&p_{s}^{i,end} = \sigma\left(\mathbf W_s^{end}\mathbf{h}_{s}^{i}+\mathbf b_s^{end}\right) \label{eq:Subject Tagger}
\end{aligned}
\end{equation}
where $p_{s}^{i,start}$ and $p_{s}^{i,end}$ represent the probabilities of  the $i$-th token  being the start token and end token of a subject respectively.  $\mathbf{W}_s^{(.)} \in \mathcal{R}^{1 \times d_h} $ is a  trainable matrix, $\mathbf{b}_s^{(.)} \in \mathcal{R}^{1}$ is a bias vector. In all  equations of this paper,  $\sigma$ denotes a sigmoid activation function.

In this study, we use a simple \emph{1/0} tagging scheme, which means a token will be assigned a \emph{1} tag if its probability exceeds a certain
threshold or a \emph{0} tag otherwise. 

\noindent \textbf{Subject-based Object Tagger} is used to extract all objects conditioned on the subjects extracted. To this end, it designs an iterative tagging structure that takes the subjects extracted  one by one and extracts all the objects for each selected subject. 

Specifically, given a selected subject,  each token in the input sentence are assigned two probabilities  to denote the possibilities of it being the start token and end token of an object that is related to this selected subject. And these two kinds of probabilities  are computed with  Eq.~\eqref{eq:Subject-based Object Tagger}.
\begin{equation}
\begin{aligned}
& \mathbf{v}_s^{s\_k} = \operatorname{maxpool}\left(\mathbf{h}_s^{s\_k\_start}, \dots, \mathbf{h}_s^{s\_k\_end}\right) \\
&{p}_{o}^{i,start} = \sigma\left(\mathbf W_{o}^{start}\left(\mathbf{h}_{o}^{i} \circ \mathbf{v}_{s}^{s\_k}\right)+\mathbf b_{o}^{start}\right)\\
&{p}_{o}^{i,end} = \sigma\left(\mathbf W_{o}^{end}\left(\mathbf{h}_{o}^{i} \circ \mathbf{v}_{s}^{s\_k}\right)+\mathbf b_{o}^{end}\right)
\label{eq:Subject-based Object Tagger}
\end{aligned}
\end{equation}
where  $\mathbf{h}_s^{s\_k\_start}, \dots, \mathbf{h}_s^{s\_k\_end}$ are the vector representations of the tokens in the $k$-th subject, so $\mathbf{v}_s^{s\_k}$ can be viewed as a  representation for the $k$-th subject. $\operatorname{maxpool}(.)$ means the \emph{max-pooling} operation. 
$p_{o}^{i,start}$ and $p_{o}^{i,end}$ are the probabilities of the $i$-th token  being the start and end tokens of an object   related to the  $k$-th subject respectively.  $\circ$  denotes a hadamard product operation.  $\mathbf{W}_o^{\left(.\right)} \in \mathcal{R}^{1 \times d_h}$ is a trainable matrix, and $\mathbf{b}_o^{\left(.\right)} \in \mathcal{R}^1$ is a bias vector.

\noindent \textbf{Cross Entropy based Losses} As mentioned above, all the extraction modules in two directions work in a multi-task learning manner. Thus, both extraction modules in each  direction have their own loss functions.
We denote the losses of above  two tagger modules  as $\mathcal{L}_{s1}$ and $\mathcal{L}_{o1}$ respectively, and both of them are  defined with a binary cross entropy based loss, as shown in Eq.~\eqref{eq:so1-loss}. 
\begin{align}
\begin{aligned}
& \operatorname{ce}\left(p,t\right)= -\left[t log p+\left(1-t\right)  log\left(1-p\right)\right] \\
& \mathcal{L}_{s1} = \frac{1}{2 \times l} \sum_{m \in\{\text{start,end}\}} \sum_{i=1}^{l} \operatorname{ce}\left(p_{s}^{i,m}, t_{s}^{{i}, {m}}\right) \\
& \mathcal{L}_{o1} = \frac{1}{2 \times l} \sum_{m \in\{\text{start,end}\}} \sum_{i=1}^{l} \operatorname{ce}\left(p_{o}^{i,m}, t_{o}^{i,m}\right) 
\label{eq:so1-loss}
\end{aligned}
\end{align}
where $\operatorname{ce}(p,t)$ is a binary cross entropy loss, $p\in(0,1)$ is the predicted probability and $t$ is the true tag, $l$ is the number of tokens in an input sentence.  

Similarly, there are two tagger losses in the \emph{o2s}  direction. We denote them  as $\mathcal{L}_{s2}$ and $\mathcal{L}_{o2}$ respectively and they are computed with the similar method as shown in  Eq.~\eqref{eq:so1-loss}.

\subsection{RE }
The proposed  framework makes \emph{BiEPE} output more  \emph{s-o} pairs, where there are many noise   pairs. This is harmful to the precision of \emph{BiRTE}. Thus,  
{RE} should have a strong classification ability. 
Here we use a biaffine model~\cite{goh1994biaffine,dozat2016deep} for the RE module. It maintains a parameter matrix for each relation, and  an entity pair will be computed with each relation-specific matrix to determine whether it possesses the corresponding relation or not. Specifically, for an entity pair ($s_k$, $o_j$), we first obtain the representation vectors $\mathbf{v}_r^{s\_k}$ and $\mathbf{v}_r^{o\_j}$ for its two entities.
Then  the possibility (denoted as  $p_r^i$) of ($s_k$, $o_j$) possessing the $i$-th relation is computed. The  process is shown in Eq.~\eqref{eq:RE}, where $\mathbf{W}_r^i \in \mathcal{R}^{(d_h+1) \times (d_h+1)} $ is the parameter matrix of the $i$-th relation. 
\begin{equation}
\begin{aligned}
& \mathbf{v}_r^{s\_k} = \operatorname{maxpool}\left( \mathbf{h}_r^{s\_k\_start}, \dots, \mathbf{h}_r^{s\_k\_end}\right) \\
& \mathbf{v}_r^{o\_j} = \operatorname{maxpool}\left(\mathbf{h}_r^{o\_j\_start}, \dots, \mathbf{h}_r^{o\_j\_end}\right) \\
& p_r^i = \sigma \left( \left[\begin{array}{l}
\mathrm{v}_r^{s\_k} \\
1
\end{array}\right]^{\top} \mathbf{W}_r^i \left[\begin{array}{l}
\mathrm{v}_r^{o\_j}  \\
1
\end{array}\right] \right) 
\label{eq:RE}
\end{aligned}
\end{equation}

Here we select the  biaffine model mainly due to its following  two  advantages. First, it maintains a matrix for each relation, which can model the characteristics of a relation accurately. Second, its probability computation mechanism makes it can accurately mine the interactions between a subject and an object. Both advantages are much helpful for improving the extraction precision. 

\noindent\textbf{RE Loss} To train the \emph{RE} component, we also define a  cross entropy based loss, as shown in Eq.~\eqref{eq:rel-loss},  where $\emph{R}$ is the predefined relation set and $\vert \emph{R} \vert$ is the number of total relations. 
\begin{equation}
\mathcal{L}_{r} = \frac{1}{\vert \emph{R} \vert} \sum_{i=1}^{\vert \emph{R} \vert} \operatorname{ce}\left(p_{r}^{i}, t_{r}^{i}\right) \label{eq:rel-loss}
\end{equation}

\subsection{Share-aware Learning Mechanism}
Totally,  there are five  extraction modules in   \emph{BiRTE}. During the multi-task learning based training, each of them will form a relative independent extraction \emph{task} with the \emph{Encoder} module.  We use the popular \emph{teacher forcing} mode to train all the \emph{tasks} except the ones that \emph{ONLY} take original sentence as input. Under this mode,  each \emph{task}  randomly selects some correct samples  as input  for training. Besides, to alleviate the \emph{exposure bias} issue~\cite{wei-etal-2020-novel}, we  merge  some randomly generated negative samples into the correct samples and use them together to train these \emph{tasks} where the \emph{teacher forcing} mode used. The negative samples can simulate the real scenario in the inference phase, which is  helpful for training a robust model. Accordingly, the mentioned \emph{exposure bias} issue is alleviated greatly.  Finally, the overall loss of \emph{BiRTE} is defined with Eq.\eqref{eq:all-loss}. 
\begin{equation}
\mathcal{L}=\mathcal{L}_{s1}+\mathcal{L}_{o1}+\mathcal{L}_{s2}+\mathcal{L}_{o2}+\mathcal{L}_{r} \label{eq:all-loss}
\end{equation}



However, we observe that the parameters in the shared  \emph{Encoder} module will receive back propagated gradients from the parameters of each extraction module. Consequently, the convergence rate of   the \emph{Encoder} module will be  much different from those in other extraction modules. This will result in a  \emph{convergence rate inconsistency issue}, which means  if we set a unified learning rate for these five extraction modules and the \emph{Encoder} module, it would be difficult for them to converge to their optimal points simultaneously. In other words, some  modules will be over-trained while others will be under-trained under a unified learning rate.  


So we propose a \emph{share-aware} learning mechanism that assigns different learning rates for different modules. The basic idea of this mechanism is that the more \emph{tasks}  a module is shared by, the smaller  learning rate it should be assigned. For example,  the \emph{Encoder} module  should be assigned a smaller learning rate than other extraction modules since it is shared by more \emph{tasks}. Specifically, the proposed  learning mechanism assigns learning rates   with  Eq.\eqref{eq:lr}.
\begin{equation}
\mathcal{\xi}_{i} = \left\{
\begin{array}{ll}
\mathcal{\xi} , \ \ \ \ \ \ \ \ \ \quad \ \ k_i=1\\
\frac{(1+\delta)}{f(k_i)} * \mathcal{\xi}, \ \ k_i >1
\end{array}
\right.
\label{eq:lr}
\end{equation}
where $\mathcal{\xi}$ is a base learning rate, $\mathcal{\xi}_{i}$ is the learning rate for the $i$-th  module and  ${k_i}$ is the number of \emph{tasks} that  the $i$-th module is shared by. For example, in \emph{BiRTE}, the corresponding  $k$ of the \emph{Encoder} module would be $5$ since this module is shared by all the five \emph{tasks}, 
while the corresponding  $k$ of the \emph{subject tagger} module in the \emph{s2o} direction would be $1$ since this module is only used by its own \emph{task}. $\delta \in [0,1]$ is a regulatory factor that is used to finely adjust the learning rate, and $f({.}) $ is a mapping function that transforms the input $k_i$ to a reasonable real value (often larger than $1$) so as  to determine the major magnitude of the learning rate.  

\section{Experiments}
\renewcommand{\arraystretch}{1} 
\begin{table}[!t] 
	\centering 
	\small
	\begin{tabular}{ccccccccc} 
		\toprule
		\multirow{2}{*}{Category}&
		\multicolumn{2}{c}{NYT}&
		\multicolumn{2}{c}{WebNLG}&
		\multicolumn{2}{c}{NYT10}&
		\multicolumn{2}{c}{NYT11}
		\\
		\cmidrule(lr){2-3}\cmidrule(lr){4-5}\cmidrule(lr){6-7}\cmidrule(lr){8-9}
		&Train&Test&Train&Test&Train&Test&Train&Test\\
		\hline 
		\textit{Normal}&37013&3266&1596&246&59396&2963&53395&368 \\
		\textit{EPO}&9782&978&227&26&5376&715&2100&0 \\
		\textit{SEO}&14735&1297&3406&457&8772&742&7365&1\\
		\hline 
		ALL&56195 &5000 &5019 &703&70339&4006&62648&369\\
		\bottomrule 
	\end{tabular} 
	\caption{Statistics of datasets. \emph{EPO} and \emph{SEO} refer to the \emph{entity pair overlapping} and \emph{single entity overlapping} respectively~\cite{zeng2018Extracting}. Note a sentence can belong to both \emph{EPO}  and \emph{SEO}.} 
	\label{tab:statistics} 
\end{table}

\subsection{Experiment Settings}
\noindent \textbf{Datasets} 
We evaluate \emph{BiRTE} on  following benchmark datasets: NYT~\cite{riedel2010Modeling}, WebNLG~\cite{gardent2017Creating}, 
NYT10~\cite{riedel2010Modeling}, and  NYT11~\cite{hoffmann2011knowledge}. 
To be fair, we follow some latest work~\cite{wei-etal-2020-novel,wang-etal-2020-tplinker,sun2021}, which  uses 
the preprocessed NYT and WebNLG datasets released by~\cite{zeng2018Extracting}, and uses the preprocessed NYT10 and NYT11 datasets released by ~\cite{takanobu2019a}.  Some  statistics of these  datasets are shown in Table~\ref{tab:statistics}.

Note that both NYT and WebNLG have two different versions according to following two annotation  standards: 1) annotating the last token of the entities, and 2) annotating the whole entity span. Different work chooses different versions of these datasets.  For convenience,  we denote the datasets based on the first standard as NYT$^*$ and WebNLG$^*$, and the datasets based on the second  standard as NYT and WebNLG. Obviously, the full annotated datasets can reveal the true performance of a model better.

Besides,  ~\cite{wei-etal-2020-novel} point out that both NYT10 and NYT11 are far less popular than either NYT or WebNLG, and they are usually used to show the generalization capability of a model because  most test sentences in them  belong to the \emph{Normal} class. Thus, for space saving, we  adopt them only in the main experiment part. 

\noindent \textbf{Evaluation Metrics} The standard micro precision, recall, and \emph{F1} score are used to evaluate the results. There are two match standards for the RTE task:  one is \emph{Partial Match} that an extracted triple is regarded as correct if the predicted relation and the head of both subject entity and object entity are  correct; and the other is \emph{Exact Match} that a  triple is regarded as  correct  only when its  entities and relation are completely matched with a correct triple. To be fair, we follow previous work~\cite{wang-etal-2020-tplinker,wei-etal-2020-novel,sun2021} and use \emph{Partial Match} on NYT$^*$ and WebNLG$^*$,  use \emph{Exact Match} on NYT and WebNLG. 


\begin{table*}[htbp]
	\begin{center}
		
		{\begin{tabular}{lccccccccccccc}
				\hline
				\multirow{3}{*}{Model}   
				&\multicolumn{6}{c}{Partial Match} & \multicolumn{6}{c}{Exact Match}& \\
				\cmidrule(lr){2-7} 
				\cmidrule(lr){8-13} 
				&\multicolumn{3}{c}{NYT$^*$} & \multicolumn{3}{c}{WebNLG$^*$}& \multicolumn{3}{c}{NYT} & \multicolumn{3}{c}{WebNLG}\\
				\cmidrule(lr){2-4} 
				\cmidrule(lr){5-7} 
				\cmidrule(lr){8-10} 
				\cmidrule(lr){11-13} 
				& Prec. & Rec. & F1 & Prec. & Rec. & F1 & Prec. & Rec. & F1 & Prec. & Rec. & F1 \\
				\hline
				ETL-Span~\cite{yu2019joint} &84.9 &72.3 &78.1 &84.0 &91.5 &87.6 &85.5 &71.7 &78.0 &84.3 &82.0 &83.1\\
				WDec~\cite{DBLP:conf/aaai/NayakN20} &-- &-- &-- &-- &-- &-- &88.1 &76.1 &81.7 &-- &-- &--\\
				RSAN~\cite{yuan2020a} &-- &-- &-- &-- &-- &-- &85.7 &83.6 &84.6 &80.5 &83.8 &82.1\\
				RIN~\cite{DBLP:conf/emnlp/SunZMML20} &87.2 &87.3 &87.3 &87.6 &87.0 &87.3 &83.9 &85.5 &84.7 &77.3 &76.8 &77.0\\
				CasRel$_{LSTM}$~\cite{wei-etal-2020-novel} &84.2 &83.0 &83.6 &86.9 &80.6 &83.7 &-- &-- &-- &-- &-- &--\\
				PMEI$_{LSTM}$~\cite{sun2021} &88.7 &86.8 &87.8 &88.7 &87.6 &88.1 &84.5 &84.0 &84.2 &78.8 &77.7 &78.2\\
				TPLinker$_{LSTM}$~\cite{wang-etal-2020-tplinker} &83.8 &83.4 &83.6 &90.8 &90.3 &90.5 &86.0 &82.0 &84.0 &{\bf 91.9} &81.6 &86.4\\
				R-BPtrNet$_{LSTM}$$^{\ddagger}$~\cite{DBLP:conf/naacl/ChenZHH21} &90.9 &91.3 &91.1 &90.7 &{\bf 94.6} &92.6 &-- &--&--&--&--&--\\
				CGT$_{UniLM}$ ~\cite{DBLP:conf/aaai/YeZDCTHC21} &{\bf 94.7} &84.2 &89.1 &92.9 &75.6& 83.4 &-- &-- &-- &-- &-- &-- \\
				CasRel$_{BERT}$~\cite{wei-etal-2020-novel} &89.7 &89.5 &89.6 &{93.4} &90.1 &91.8 &89.8$^{\star}$ &88.2$^{\star}$ &89.0$^{\star}$  &88.3$^{\star}$ &84.6$^{\star}$ &86.4$^{\star}$\\
				PMEI$_{BERT}$~\cite{sun2021} &90.5 &89.8 &90.1 &91.0 &92.9 &92.0 &88.4 &88.9 &88.7 &80.8 &82.8 &81.8\\
				TPLinker$_{BERT}$~\cite{wang-etal-2020-tplinker} &91.3 &92.5 &91.9 &91.8 &92.0 &91.9 &91.4 &92.6 &92.0 &88.9 &84.5 &86.7\\		
				StereoRel$_{BERT}$~\cite{DBLP:conf/acl/TianJHL20} &92.0 &92.3 &92.2 &91.6 &92.6 &92.1 &92.0 &92.3 &92.2  &--&--&-- \\
				PRGC$_{BERT}$ ~\cite{DBLP:conf/acl/ZhengWCYZZZQMZ20} &93.3 &91.9 &92.6 &94.0 &92.1 &93.0 &{\bf 93.5} &91.9 &92.7 &89.9 &87.2 &88.5\\
				
				R-BPtrNet$_{BERT}$$^{\ddagger}$~\cite{DBLP:conf/naacl/ChenZHH21} &92.7 &92.5 &92.6 &\textbf{93.7} &92.8 &93.3 &-- &--&--&--&--&--\\
				\hline
				\textbf{BiRTE}$_{LSTM}$ &86.5 &89.0 &87.7 &90.5 &91.6 &91.0 &86.4 &87.1 &86.7 &85.2 &87.3 &86.2\\
				\textbf{BiRTE}$_{BERT}$ &{92.2} &\textbf{93.8} &\textbf{93.0} &93.2 &{94.0} &\textbf{93.6} &{91.9} &\textbf{93.7} &\textbf{92.8} &{89.0} &\textbf{89.5} &\textbf{89.3}\\
		\end{tabular}}
		{\begin{tabular}{lccccccccccccc}
				\hline
				\multirow{3}{*}{Model}   
				&\multicolumn{6}{c}{Partial Match} & \multicolumn{6}{c}{Exact Match}& \\
				\cmidrule(lr){2-7} 
				\cmidrule(lr){8-13} 
				&\multicolumn{3}{c}{NYT10} & \multicolumn{3}{c}{NYT11}& \multicolumn{3}{c}{NYT10} & \multicolumn{3}{c}{NYT11} \\
				\cmidrule(lr){2-4} 
				\cmidrule(lr){5-7} 
				\cmidrule(lr){8-10} 
				\cmidrule(lr){11-13} 
				
				& Prec. & Rec. & F1 & Prec. & Rec. & F1 & Prec. & Rec. & F1 & Prec. & Rec. & F1 \\
				\hline
				PMEI$_{LSTM}$~\cite{sun2021}  &79.1&67.2&72.6&56.0&58.6&57.2&75.4&65.8&70.2&{\bf 55.3}&57.8&56.5\\
				CasRel$_{BERT}$~\cite{wei-etal-2020-novel} &77.7 &68.8 &73.0 &50.1 &58.4 &53.9&76.8$^\star$&68.0$^\star$&72.1$^\star$&49.1$^\star$&56.4$^\star$&52.5$^\star$\\
				
				StereoRel$_{BERT}$~\cite{DBLP:conf/acl/TianJHL20} &80.0 &67.4 &73.2  &53.8 &55.4 &54.6 &--&--&-- &--&--&-- \\
				
				PMEI$_{BERT}$~\cite{sun2021} &79.1 &70.4 &74.5 &55.8 &59.7 &57.7&77.3 &69.7 &73.3 &54.9 &58.9 &56.8 \\

				
				TPLinker$_{BERT}$~\cite{wang-etal-2020-tplinker} &78.9$^\star$&71.1$^\star$&74.8$^\star$&55.9$^\star$&60.2$^\star$&58.0$^\star$&78.5$^\star$&68.8$^\star$&73.4$^\star$
				&54.8$^\star$&59.3$^\star$&57.0$^\star$\\		
				\hline
				\textbf{BiRTE}$_{LSTM}$ &79.0&68.8&73.5&55.1&60.4&57.6&76.1&67.4&71.5&54.1&60.5&57.1\\
				\textbf{BiRTE}$_{BERT}$ &\textbf{80.6}&\textbf{71.8}&\textbf{76.0}&\textbf{56.4}&\textbf{62.0}&\textbf{59.1}&\textbf{80.1}&\textbf{71.4}&\textbf{75.5}&{55.0}&\textbf{61.2}&\textbf{57.9}\\
				\hline
		\end{tabular}}
		{
			\caption{Main experiments. Note \emph{CGT} uses  \emph{UniLM}~\cite{dong2019unified}. $^{\ddagger}$: R-BPtrNet uses extra entity type features while all other models not. }
			\label{table:main1}
		} 
	\end{center}
\end{table*}

\noindent \textbf{Implementation Details}  AdamW~\cite{KingmaB14} is used to train \emph{BiRTE}. The threshold for judging whether there a subject, an object, or a relation is set to $0.5$. In Eq.\eqref{eq:lr}, $\xi$ is set to $1.5e^{-4}$, the regulatory factor $\delta$ is set to $0$,  and the mapping function $f(.)$ is defined as an identity function. The  batch size is set to 18 on  NYT, NYT$^*$, NYT10 and NYT11, and  is set to 6 on WebNLG and WebNLG$^*$.   All involved   hyperparameters are determined based on the results on the development sets. Other parameters are randomly initialized. 
In  experiments, all the involved \emph{BERT} model refers to \emph{BERT (base)}. On all datasets, we run our model 5 times and the averaged results are taken as the final reported results. 

\noindent \textbf{Baselines} Following   strong state-of-the-art  models are taken as baselines, including: 
	  \emph{ETL-Span}~\cite{yu2019joint}, \emph{WDec}~\cite{DBLP:conf/aaai/NayakN20},
	   \emph{RSAN}~\cite{yuan2020a},
	   \emph{RIN}~\cite{DBLP:conf/emnlp/SunZMML20},  \emph{CasRel}~\cite{wei-etal-2020-novel},   \emph{TPLinker}~\cite{wang-etal-2020-tplinker},
	   \emph{StereoRel}~\cite{DBLP:conf/acl/TianJHL20}, \emph{PRGC}~\cite{DBLP:conf/acl/ZhengWCYZZZQMZ20}, \emph{R-BPtrNet}~\cite{DBLP:conf/naacl/ChenZHH21}, \emph{PMEI}~\cite{sun2021}, and \emph{CGT}~\cite{DBLP:conf/aaai/YeZDCTHC21}.  
Most  results  of these baselines  are copied from their original papers directly. Moreover, following previous work~\cite{wei-etal-2020-novel,sun2021,wang-etal-2020-tplinker,DBLP:conf/naacl/ChenZHH21}, we also implement a \emph{BiLSTM}-encoder version of \emph{BiRTE} where   300-dimensional GloVe embeddings~\cite{pennington2014Glove} and 2-layer stacked BiLSTM are used. 
Some baselines did not report their results on some   datasets. In  such case, we report the best results we obtained (marked by  $^\star$) by running the source code (if available).  
But if a baseline did not report the results of its \emph{BiLSTM}-encoder version, we would not obtain these results  even if the source code is available:   because it needs to modify the provided source code in such case, which will increase the concern of whether such modification is correct and whether the obtained results are objective.

\subsection{Experimental Results}
\textbf{Main Results} The main results are shown in  Table~\ref{table:main1}. On all datasets, \emph{BiRTE}  achieves almost all  the best results  in term of \emph{F1}  when  compared with the models that use the same kind of encoder (\emph{BERT} or \emph{BiLSTM}). When considering the complete version of each model where \emph{BERT}  used,   \emph{BiRTE} works much better  than all the compared baselines: it achieves the best results on almost all datasets in term of recall and \emph{F1}.   \emph{BiRTE} achieves  slightly poor but still much competitive precision results. 
This is in line with our previous analyses that  some noise pairs are extracted by the bidirectional  framework, which is  harmful to  precision. However, the proposed framework brings much more benefits on recall, which makes \emph{BiRTE} achieves  much higher \emph{F1} scores. Another interesting observation is that \emph{BiRTE} achieves far better results than 
 \emph{CasRel},  which proves the correctness of our motivation. 
 

Besides, \emph{BiRTE} achieves better \emph{F1} results  on all the full annotated datasets. This is very meaningful because it indicates that \emph{BiRTE} will  perform well when  deployed  in real scenarios where both the \emph{full annotation} standard and   the  \emph{exact match} standard are usually required.   \emph{BiRTE} also achieves much better results than all the compared baselines on both NYT10 and NYT11, which indicates that \emph{BiRTE} has a good generalization capability.
 
In subsequent sections, we evaluate \emph{BiRTE}  from diverse aspects, and all the results are obtained when the BERT-based encoder used.  

\noindent\textbf{Evaluations on Complex Sentences} Here we  evaluate \emph{BiRTE}'s  ability for extracting triples from sentences that contain overlapping triples and multiple triples.  This ability is widely  discussed in existing  models, and is an important metric to evaluate the robustness of a model. For fair comparison,    we follow the  settings of some previous best models~\cite{wei-etal-2020-novel,wang-etal-2020-tplinker,DBLP:conf/acl/ZhengWCYZZZQMZ20,DBLP:conf/naacl/ChenZHH21}, which are: (i) classifying sentences according to the degree of overlapping and the number of triples contained in a sentence, and (ii) conducting experiments on different subsets of NYT$^*$ and WebNLG$^*$.



The results are shown in Table~\ref{table:f1_on_split}. We can see  that \emph{BiRTE} has great  superiority for handling complex sentences. On both datasets, it  achieves   much better results than the compared baselines on most cases. 
Moreover, \emph{BiRTE} achieves  more performance improvement when handling the sentences of \emph{SEO} class. This is mainly because that   a single entity  in a \emph{SEO} sentence may associate with multiple triples, thus the existing models (even including the non-tagging based models like \emph{TPLinker}) are more likely to suffer from the \emph{ground entity extraction failure} issue on the \emph{SEO} sentences than on other types of sentences: once the extraction of an entity in some \emph{SEO} triples is failed, all the associated triples of this entity  would not be extracted either. 
But the bidirectional  framework in \emph{BiRTE} can effectively overcome such deficiency and the mentioned issue almost has no effect on it when handling the \emph{SEO} sentences. This is also the reason why \emph{BiRTE}  performs  well on sentences that contain  multiple triples. Note \emph{R-BPtrNet}~\cite{DBLP:conf/naacl/ChenZHH21} also achieves very competitive results, which is partly because it uses extra entity type knowledge.  
	\begin{table*}[t]
	\begin{center}
		
		\setlength{\tabcolsep}{0.5 mm}{\begin{tabular}{lccccccccccccccccc}
				\hline
				\multirow{2}{*}{Model} & \multicolumn{8}{c}{NYT$^*$}& \multicolumn{9}{c}{WebNLG$^*$} \\
				& Normal & SEO & EPO & T = 1 & T = 2 & T = 3 & T = 4 & T $\geq$ 5 && Normal & SEO & EPO & T = 1 & T = 2 & T = 3 & T = 4 & T $\geq$ 5 \\
				\hline
				CasRel$_{BERT}$~\cite{wei-etal-2020-novel}  & 87.3 & 91.4 & 92.0 & 88.2 & 90.3 & 91.9 & 94.2 & 83.7 &&89.4 & 92.2 & 94.7 &89.3 &90.8 & 94.2 & 92.4 & 90.9  \\
				TPLinker$_{BERT}$~\cite{wang-etal-2020-tplinker}  & {90.1} & {93.4} & {94.0} & {90.0} & {92.8} & {93.1} & {96.1} & {90.0} && 87.9 & {92.5} & {95.3} & 88.0 & 90.1 & {94.6} & {93.3} & {91.6}\\ 
				PRGC$_{BERT}$~\cite{DBLP:conf/acl/ZhengWCYZZZQMZ20} &91.0 &94.0 &94.5 &91.1 &93.0 &93.5 &95.5 &\textbf{93.0} & &90.4 &93.6 &95.9 &89.9 &91.6 &95.0 &94.8 &92.8 \\
				R-BPtrNet$_{BERT}$~\cite{DBLP:conf/naacl/ChenZHH21}  & {90.4} & {94.4} & \textbf{95.2} & {89.5} & {93.1} & {93.5} & \textbf{96.7} & {91.3} && 89.5 & {93.9} & \textbf{96.1} & 88.5 & 91.4 & \textbf{96.2} & \textbf{94.9} & \textbf{94.2}\\
				
				\hline
				BiRTE$_{BERT}$&\textbf{91.4}&\textbf{94.7}&{94.2}&\textbf{91.5}&\textbf{93.7}&\textbf{93.9}&{95.8}&{92.1}&&\textbf{90.1}&\textbf{95.9}&94.3&\textbf{90.2}&\textbf{92.9}&{95.7}&{94.6}&{92.0}\\
				\hline
		\end{tabular}}
		\caption{F1 scores on sentences with different overlapping pattern and different triplet number. Results of \emph{CasRel} are copied from \emph{TPLinker} directly. ``T'' is the number of triples contained in a sentence. 
		}
		\label{table:f1_on_split}
	\end{center}
\end{table*}
\begin{table*}[htbp]
	\begin{center}
		
		\setlength{\tabcolsep}{1.5mm}{\begin{tabular}{lccccccccccccc}
				\hline
				\multirow{3}{*}{Model}   
				&\multicolumn{6}{c}{Partial Match} & \multicolumn{6}{c}{Exact Match}& \\
				\cmidrule(lr){2-7} 
				\cmidrule(lr){8-13} 
				&\multicolumn{3}{c}{NYT$^*$} & \multicolumn{3}{c}{WebNLG$^*$}& \multicolumn{3}{c}{NYT} & \multicolumn{3}{c}{WebNLG}\\
				\cmidrule(lr){2-4} 
				\cmidrule(lr){5-7} 
				\cmidrule(lr){8-10} 
				\cmidrule(lr){11-13} 
				& Prec. & Rec. & F1 & Prec. & Rec. & F1 & Prec. & Rec. & F1 & Prec. & Rec. & F1 \\
				
				\hline
				BiRTE$_{BERT}$ &\textbf{92.2} &\textbf{93.8} &\textbf{93.0} &\textbf{93.2} &\textbf{94.0} &\textbf{93.6} &\textbf{91.9} &93.7 &\textbf{92.8} &\textbf{89.0} &\textbf{89.5} &\textbf{89.3}\\
				\hline

				BiRTE$_{s2o}$&91.5&91.3&91.4&92.0&90.4&91.2&91.5 &91.0 &91.2&88.3&87.0&87.6\\
				BiRTE$_{o2s}$&91.4&91.0&91.2&91.8&90.5&91.1&91.5&90.8&91.1&88.5&87.5&88.0\\
				BiRTE$_{Fine Pipeline}$&90.4&91.2&90.8&91.0&91.6&91.3&89.7&90.1&89.9&84.0&85.6&84.8\\
				BiRTE$_{Coarse Pipeline}$&90.9&92.3&91.6&91.9&92.1&92.0&90.5&91.0&90.7&85.7&87.3&86.5\\
				
				\hline
				
				BiRTE$_{OneLr}$&91.0&92.4&91.7&92.5&93.6&93.0&91.2&91.8&91.5 &88.1&89.0&88.5\\
				
				BiRTE$_{uif}$&91.6&92.9&92.2&92.7&93.8&93.2&91.3&92.5&91.9 &88.8&88.6&88.7\\
				BiRTE$_{tru}$&92.1&93.4&92.7&93.2&93.8&93.5&91.5&93.2&92.3 &88.9&89.3&89.1\\
				\hline
				BiRTE$_{BIO}$&92.1&93.7&92.9&93.0&93.9&93.4&\textbf{91.9}&{93.8}&\textbf{92.8}&88.8&\textbf{89.5}&89.1\\
				BiRTE$_{2step}$&89.5&92.3&90.9&89.9&91.9&90.9&89.0&91.5&90.2&84.7&87.6&86.1\\	
				\hline			
				BiRTE$_{Li}$&91.0&93.6&92.3&91.6&92.9&92.2&90.5&\textbf{93.9}&92.2 &87.2&89.3&88.2\\
				
				
				
				\hline
		\end{tabular}}
		{
			\caption{Results of detailed evaluations. }
			\label{table:abl1}
		} 
	\end{center}
\end{table*}

\begin{table}[t]
	\centering
	{ 
		
		\begin{tabular}{llllll}
			\hline
			\multirow{1}{*}{Models}& \multirow{1}{*}{Direction}&\multicolumn{1}{l}{NYT$^*$}&\multicolumn{1}{l}{WebNLG$^*$}&\multicolumn{1}{l}{NYT}&\multicolumn{1}{l}{WebNLG}\\
			\hline
			\multirow{2}{*}{BiRTE} & s2o  &95.0&95.3 &94.2   &91.0\\  
			& o2s    &94.8&95.6  &93.9   &91.1\\
			\hline   
			\multirow{1}{*}{BiRTE$_{s2o}$}&s2o  &93.6&92.6&93.1 &89.3 \\
			\multirow{1}{*}{BiRTE$_{o2s}$}   
			& o2s  &93.2&92.8  &92.8&89.5\\
			
			\hline%
		\end{tabular}
	}
	\caption{F1 results of the ground entity extraction.}
	\label{tab:BiEPE}
\end{table}

\begin{table}[t]
	\centering
	{
		\begin{tabular}{llllll}
			\hline
			\multirow{1}{*}{Models}&\multicolumn{1}{l}{NYT$^*$}&\multicolumn{1}{l}{WebNLG$^*$}&\multicolumn{1}{l}{NYT}&\multicolumn{1}{l}{WebNLG}\\
			\hline
			\multirow{1}{*}{ETL-Span}&54.3&56.1&56.8&60.2\\
			\multirow{1}{*}{CasRel}&49.7&48.5&55.7&51.8\\
			\multirow{1}{*}{BiRTE$_{s2o}$}&55.2 &39.6 &56.0&42.8\\
			\multirow{1}{*}{BiRTE$_{o2s}$} &53.5&51.2 &54.8&52.2\\
			\hline
			\multirow{1}{*}{BiRTE}&9.7&5.4 &11.0&9.3\\
			
			\hline%
		\end{tabular}
	}
	\caption{Proportions (\%) of triples that are not  extracted due to the \emph{ground entity extraction failure} issue. }
	\label{tab:error}
\end{table}

\noindent\textbf{Detailed Evaluations} 
Here we make three kinds of detailed evaluations on \emph{BiRTE}, and the results are shown in Table  \ref{table:abl1}.

\textbf{First}, we evaluate the contributions of the proposed bidirectional extraction framework from following four aspects. 


\textbf{(1)} We evaluate whether  the proposed bidirectional extraction framework  is superior to the  unidirectional extraction frameworks. To this end,  we implement following two variants of \emph{BiRTE}:  (i) BiRTE$_{s2o}$, a variant that only uses the \emph{s2o} direction to extract entity pairs, based on which the triples are extracted; (ii) BiRTE$_{o2s}$, a variant  that only uses the \emph{o2s} direction to extract entity pairs, based on which the triples are extracted.  
Results show that    the performance of both variants drops on all datasets, which shows the superiority of the proposed bidirectional framework. Especially, both variants achieve   lower recalls, which indicates again that  the unidirectional extraction framework based models are  sensitive to the \emph{ground entity extraction failure} issue. While in \emph{BiRTE}, the two directions' \emph{s-o} pair extractions can boost each other, so the mentioned issue is alleviated greatly, which is much helpful for recall. 

\textbf{(2)} We evaluate whether the proposed bidirectional extraction framework does be helpful for extracting better ground entities than the unidirectional frameworks.  To this end, we compare the ground entities' extraction results  between \emph{BiRTE}, BiRTE$_{s2o}$, and BiRTE$_{o2s}$. The  results   are shown in  Table \ref{tab:BiEPE}. 
We can see that  in each  direction, \emph{BiRTE} achieves much better extraction results than its  variant of the same direction.  
This is mainly because that with the help of the multi-task learning mechanism,   the ground entity extractions of two directions  can boost each other by the explicitly injected context features through the shared  \emph{Encoder} component, which is much helpful for the extraction results of each direction.  

\textbf{(3)} We compare  the proportion of the triples that are not extracted  due to the \emph{ground entity extraction failure} issue between  \emph{BiRTE} and other tagging based methods that take an unidirectional extraction framework. This proportion can quantitatively  demonstrate both the severity caused by the mentioned issue and  the practical  effect of the proposed bidirectional extraction framework. The  results are shown in Table \ref{tab:error}. 

We can see that  for all the unidirectional extraction framework based models, almost half of the failed extracted triples  are caused by the  mentioned \emph{ground entity extraction failure} issue. While for \emph{BiRTE}, this proportion drops sharply. These results show that  the harm of the mentioned issue is eliminated greatly by the proposed bidirectional   framework. 






\textbf{(4)} We evaluate whether a simple combination of two paralleled extraction components can also performs well like the proposed framework.  To this end, we implement following two variants of  \emph{BiRTE}, both of which are \emph{pipeline}-based models.
(i) BiRTE$_{FinePipeline}$, a model that splits \emph{Subject Tagger}, \emph{Object Tagger}, \emph{Subject-based Object Tagger}, \emph{Object-based Subject Tagger}, and \emph{RE} into five separated models that do not share  the \emph{Encoder} component; and 
(ii) BiRTE$_{CoarsePipeline}$, a model that splits \emph{BiEPE}  and \emph{RE}  into two separated models that do not share the  \emph{Encoder} component. 

Results show that  the performance of both variants drops sharply on all datasets, which indicates that the proposed extraction framework should NOT be viewed as a simple  combination of two  individual extraction components. In fact, under the multi-task learning mechanism, the \emph{Encoder}-share structure in our framework enables different modules complement and boost each other, which is much helpful for the performance of the whole RTE task. For example,  in each direction, either  \emph{Subject Tagger}  or  \emph{Object Tagger}  will push  parameters in  \emph{Encoder}  to be updated in the way that is beneficial for its own extraction. As these two  taggers are performed alternately in the multi-task learning manner,  features that are beneficial for the \emph{subject extraction} are injected into the parameters of  \emph{Encoder}  by the back propagated gradients, based on which the \emph{object extraction} is performed, and vice versa. Thus,  the \emph{subject-related} features are implicitly used for \emph{object extraction}, which makes two taggers complement and boost each other.. 
Besides,  both  variants have a greater \emph{F1} degradation on WebNLG$^*$ and WebNLG than that  on other two datasets. This is mainly because  WebNLG is a  sparse dataset for it contains  a smaller  number of training samples but  a larger number of relations. Thus on WebNLG, the scarcity of training samples can be effectively compensated by the proposed  framework by making the correlated modules boost each other. 

\begin{table*}[htbp]
	\begin{center}
		
		\setlength{\tabcolsep}{1.5mm}{\begin{tabular}{lccccccccccccc}
				\hline
				\multirow{3}{*}{Model}   
				&\multicolumn{6}{c}{Partial Match} & \multicolumn{6}{c}{Exact Match}& \\
				\cmidrule(lr){2-7} 
				\cmidrule(lr){8-13} 
				&\multicolumn{3}{c}{NYT$^*$} & \multicolumn{3}{c}{WebNLG$^*$}& \multicolumn{3}{c}{NYT} & \multicolumn{3}{c}{WebNLG}\\
				\cmidrule(lr){2-4} 
				\cmidrule(lr){5-7} 
				\cmidrule(lr){8-10} 
				\cmidrule(lr){11-13} 
				& Prec. & Rec. & F1 & Prec. & Rec. & F1 & Prec. & Rec. & F1 & Prec. & Rec. & F1 \\
				\hline
				ETL-Span$_{BiDir}$&84.6&73.5($\uparrow$)&78.7($\uparrow$)&83.3&92.0($\uparrow$)&87.4&85.2&73.0($\uparrow$)&78.6($\uparrow$) &83.5 &83.1($\uparrow$) &83.3($\uparrow$)\\
				
				CasRel$_{BiDir}$&89.0&91.1($\uparrow$)&90.0($\uparrow$)&92.6&91.2($\uparrow$)&91.9($\uparrow$)&89.0&90.1($\uparrow$)&89.5($\uparrow$)&87.1&85.1($\uparrow$)&86.1\\
				
				ETL-Span$_{SaLr}$&85.3($\uparrow$)&73.0($\uparrow$)&78.7($\uparrow$)&84.3($\uparrow$)&91.7($\uparrow$)&87.8($\uparrow$)&86.2($\uparrow$)&72.3($\uparrow$)&78.6($\uparrow$) &83.0 &84.6($\uparrow$) &83.8($\uparrow$) \\
				
				CasRel$_{SaLr}$&90.1($\uparrow$)&89.9($\uparrow$)&90.0($\uparrow$)&93.5($\uparrow$)&90.5($\uparrow$)&92.0($\uparrow$)&90.1($\uparrow$)&89.1($\uparrow$)&89.6($\uparrow$) &87.9&87.1($\uparrow$)&87.5($\uparrow$) \\
				\hline
		\end{tabular}}
		{
			\caption{Adaptability evaluations. ``$\uparrow$'' denotes the performance is increased.  }
			\label{table:adapt}
		} 
	\end{center}
\end{table*}

\textbf{Second}, we evaluate the contributions of the proposed share-aware learning mechanism from following two aspects. 

\textbf{(1)} We evaluate the performance difference between using and without using  the proposed learning mechanism.  To this end, we implement BiRTE$_{OneLr}$, a variant of \emph{BiRTE} that  uses an identical learning rate.
From the comparison results we can see  that   the performance of  {BiRTE}$_{OneLr}$ drops obviously on  all  datasets, which indicate: (i)  the \emph{convergence rate inconsistency issue} does exist in the models where contain some shared modules; and (ii) the proposed  learning mechanism is effective for addressing this issue. 

\textbf{(2)} We evaluate the influence of the  mapping function  in the proposed  learning mechanism. To this end, we explore following two kinds of mapping functions.  (i) An  uniform increasing function  $f(k_i) = 1+2(n_i-1)k_i/(n-1) \in [1, 1+2k_i] $, where $n$ is the total number of epochs, and $n_i$ is the current epoch number. (ii) A truncated function  $f(k_i) = min (k_i, 1+2(n_i-1)k_i/(n-1)) \in [1, k_i]$.   We denote the variants of \emph{BiRTE} that use  these two mapping functions as BiRTE$_{uif}$ and BiRTE$_{tru}$ respectively. 
Results show that the mapping function  has an obvious influence on the performance.  But  all the models that use the proposed learning mechanism achieve significant  better results than BiRTE$_{OneLr}$, which confirms again  the proposed learning mechanism is  effective.  Note   the mapping function selection is still an open issue and calls for further research. 

\textbf{Third}, we conduct experiments to answer following two issues to further demonstrate the effectiveness of \emph{BiRTE}. 




\textbf{(1)} \emph{BiRTE} uses the \emph{1/0} tagging scheme in \emph{BiEPE}. However, there are other widely used  schemes like  \emph{BIO}, which can provide more richful label information than the \emph{1/0}  scheme. Thus, there is an issue: whether it would be better when the \emph{BIO} scheme  used?

To answer this issue, we implement BiRTE$_{BIO}$, a variant that uses the  \emph{BIO} scheme.  We can see  that  the performance of BiRTE$_{BIO}$ drops slightly on most of cases except on NYT where it achieves close results with \emph{BiRTE}.  
In fact, there are two advantages in the \emph{1/0} scheme. First, its labels can realize the roles of most labels in the \emph{BIO} scheme. Second, it reduces the complexity of a tagging model because with this simpler scheme, for each token, a  model only needs to distinguish whether it is an entity token or not, other than to distinguish whether this token is a \emph{beginning} or  \emph{inside} token of an entity, or not an entity token. Obviously, this simplification   reduces the risk of introducing tagging errors. 


\textbf{(2)} The \emph{ground entity extraction failure} issue can also be solved by a simple strategy that  firstly extracting all entities without distinguishing subject and object, and then using the \emph{RE} module to classify all entity pairs. 
Accordingly, there would be an issue: whether a simpler \emph{2-step extraction strategy} would work better? 

To answer this issue, we implement BiRTE$_{2step}$, a \emph{2-step extraction} based model. 
Results show that the performance of BiRTE$_{2step}$ drops sharply  compared with \emph{BiRTE}. Especially, the degradation of its precision is far larger than that of its recall on all datasets. This indicates that by considering all combinations of entity pairs,  the \emph{ground entity extraction failure} issue is alleviated to some extent. However,  among these combinations,  there are lots of noise pairs that have no any relation, which results in a more significant degradation in  precision. Consequently, its  F1 score drops.  These results indicate BiRTE$_{2step}$ is not a good choice to address the mentioned issue because it often results in far larger degradation in precision, which neutralizes the benefits from the improvement of recall. 

\noindent\textbf{Adaptability Evaluations}
In fact, both the proposed bidirectional extraction framework and the proposed share-aware learning mechanism  are adaptive and  can  be easily transplanted to other models. Here we evaluate their  adaptabilities by transplanting them to  \emph{CasRel} and \emph{ETL-Span}. Both these selected two models are state-of-the-art tagging  based methods and have a shared \emph{Encoder}. 

Specifically,  we denote the new models that use  the proposed bidirectional extraction framework   as CasRel${}_{BiDir}$ and ETL-Span${}_{BiDir}$ respectively. Both \emph{CasRel} and \emph{ETL-Span} first extract subjects, then extract  objects and relations simultaneously. Here in their new  variants,  we  simply merge the triples extracted  from two directions as  final outputted triples. We denote the new models that use the proposed share-aware learning mechanism  as ETL-Span${}_{SaLr}$ and CasRel${}_{SaLr}$ respectively.
The  results are shown in Table \ref{table:adapt}.

We can see that on almost all datasets, both CasRel${}_{BiDir}$ and ETL-Span${}_{BiDir}$ achieve better  performance than their original versions in term of \emph{F1} and \emph{recall}. These results  further confirm  that the bidirectional extraction framework can well address the \emph{ground entity extraction failure} issue, which is much helpful for recall. These two new models' \emph{precision} scores are lower than their original versions, this is because that there are more noise introduced by the bidirectional extraction framework, thus a stronger relation classification model  is required. For example, when replacing the biaffine model with  a common linear classification model that takes the concatenation of two entities' representations as input, the performance of \emph{BiRTE} ({BiRTE}$_{Li}$ in Table~\ref{table:abl1}) drops accordingly. 
We can also see that when the proposed share-aware learning mechanism used,  both CasRel${}_{SaLr}$ and ETL-Span${}_{SaLr}$ achieve better  results than their original versions on both datasets under almost all evaluation metrics, even slightly better than CasRel${}_{BiDir}$ and ETL-Span${}_{BiDir}$.

\section{Conclusions}
In this paper, we propose a  simple but effective RTE model.  There are two  main contributions in our work. 
First, we observe the \emph{ground entity extraction failure} issue existed in existing tagging based RTE methods, and propose a bidirectional extraction framework to address it.  
Second, we observe the \emph{ convergence rate inconsistency} issue existed in the share structures,  and propose a \emph{share-aware} learning mechanism to address it. 
We conduct  extensive experiments on multiple  benchmark datasets to evaluate the proposed model from diverse aspects. Experimental results show  
that the  two proposed mechanisms   are effective and adaptive, and they help our model achieve state-of-the-art results on all of these benchmark datasets. 

\begin{acks}
	This work is supported by the National Natural Science Foundation of China (No.61572120 and No.U1708261), the Fundamental Research Funds for the Central Universities (No.N181602013 and No.N2016006), Shenyang Medical Imaging Processing Engineering Technology Research Center (17-134-8-00),  Ten Thousand Talent Program (No.ZX20200035), and Liaoning Distinguished Professor (No.XLYC1902057).
\end{acks}
\bibliographystyle{ACM-Reference-Format}
\balance
\bibliography{sample-base}


\end{document}